\PassOptionsToPackage{monochrome}{xcolor}
\documentclass[times, final, 10pt]{elsarticle}

\usepackage{hyperref}
\usepackage{url}

\usepackage{times}
\usepackage{amssymb, amsmath}
\usepackage{graphicx}
\usepackage{float}
\usepackage{booktabs}

\usepackage{epstopdf}
\usepackage[table]{xcolor}
\usepackage{amsmath}
\usepackage{numprint}
\usepackage{subcaption}
\usepackage{caption}
\usepackage{multirow}

\usepackage{array}
\usepackage{calc}
\usepackage{calc}
\usepackage{tikz}

\usetikzlibrary{positioning}
\usepackage{algorithmic}
\usepackage{algorithm}
\usepackage{xcolor}
\usepackage{float}

\usepackage[export]{adjustbox}
\usepackage{multirow}
\usepackage{longtable}
\newcolumntype{M}[1]{>{\centering\arraybackslash}m{#1}}

\usepackage{tabularx}
\usepackage{hyperref}
\usepackage{cleveref}
\usepackage{colortbl}  
\usepackage[dvipsnames,table]{xcolor}
\usepackage{ragged2e,makecell,array}

\usepackage{pdflscape}
\usepackage{xltabular}
\usepackage{setspace}
\newcolumntype{L}[1]{>{\raggedright\arraybackslash}p{#1}}
\definecolor{rulegray}{HTML}{CCCCCC}
\newcommand{\legendbox}[2]{%
  \begingroup\setlength{\fboxsep}{1pt}%
  \colorbox{#1}{\strut\hspace{8pt}}\kern6pt\textcolor{black}{#2}\endgroup}

\newcolumntype{P}[1]{>{\RaggedRight\arraybackslash}p{#1}}
\newcolumntype{Y}{>{\RaggedRight\arraybackslash}X}
\definecolor{keyA}{HTML}{E6F4EA} 
\newcommand{\TableFont}{\fontsize{8}{9.6}\selectfont} 

\usepackage{etoolbox}

\AtBeginEnvironment{table}{\TableFont}
\AtBeginEnvironment{table*}{\TableFont}
\AtBeginEnvironment{xltabular}{\TableFont}
\AtBeginEnvironment{longtable}{\TableFont}
\usepackage{tabularx}
\usepackage{booktabs}
\usepackage{makecell}
\usepackage{array}
\usepackage{setspace}

\newcolumntype{Z}{>{\raggedright\arraybackslash}X}
\newcolumntype{C}[1]{>{\centering\arraybackslash}p{#1}}

\renewcommand{\rowcolor}[2][]{}

\begin{document}

\begin{frontmatter}

\title{Gaussian Splatting for Real-Time Scene Reconstruction: Methods, Systems, and Trade-offs}
\author{Ashim Dahal}
\author{Rabab Abdelfattah}
\author{Nick Rahimi}

\affiliation{organization={School of Computing and Computer Engineering, University of Southern Mississippi},
            city={Hattiesburg},
            state={MS},
            postcode={39401},
            country={USA}}
\begin{abstract}
3D Gaussian Splatting (3DGS) allows for high-quality real-time novel-view synthesis.
{\color{red}
Since its inception in 2023, its explicit and highly configurable scene representation has been extended toward sparse-view generation, dynamic reconstruction, efficiency and compression, streaming, and simultaneous localization and mapping (SLAM).
Although Gaussian Splatting can provide real-time rendering, rendering an existing representation is distinct from constructing, updating, or maintaining the scene itself.
}
In this survey, we synthesize current literature, categorize them and provide a connection of current approaches and tools towards an end-to-end pipeline.
{\color{red}
Explicitly, we categorize GS scene reconstruction into five overlapping families: single-shot generation, dynamic representations, efficiency and compression, streaming and immersive systems, and Gaussian Splatting SLAM.
We distinguish render-only throughput from task-specific scene generation, update, tracking, mapping, and optimization costs, and combine controlled sparse-view benchmarking with native evaluations to identify cross-family quality, latency, memory, and update-cost trade-offs.
}
\end{abstract}

\begin{keyword}
Gaussian Splatting \sep
3D Scene Representation \sep
Novel View Synthesis \sep
Neural Rendering \sep
Multi-View Geometry
\end{keyword}

\end{frontmatter}

\section{Introduction}

{\color{red}
Kerbl et al.~\cite{3dgs_og} introduced 3D Gaussian Splatting (3DGS) as an explicit scene representation capable of high-quality interactive novel-view rendering. Subsequent work rapidly extended this formulation beyond static per-scene reconstruction. Deformable 3DGS~\cite{Yang2023Deformable3G} and 4DGaussians~\cite{4dgs_og} introduced temporal deformation and spatiotemporal representations, while later methods explored feed-forward sparse-view generation~\cite{Charatan2023PixelSplat3G,Chen2024MVSplatE3}, efficiency and compression, streaming, and Gaussian-based SLAM. These developments increasingly shift the systems bottleneck from rendering an existing representation toward constructing, updating, transmitting, or maintaining the Gaussian scene.
}
{\color{red}
This distinction is important for applications such as free-viewpoint video, immersive media, online mapping, and robotics. In these settings, interactive rendering alone is insufficient when scene construction, temporal updates, or map maintenance remain substantially slower than rasterization.
}

{\color{red}
Several surveys have reviewed 3DGS from broader perspectives. Wu et al.~\cite{wu2024recent3dgs} organize the field by 3D reconstruction, editing, and downstream applications; Fei et al.~\cite{fei2025newera} use taxonomies based on characteristics and applications; and Bao et al.~\cite{bao2025gaussiansurvey} examine optimization, applications, extensions, and recurring technical modules. More focused surveys address compression and compaction~\cite{bagdasarian2025gszip} and robotics~\cite{zhu2024roboticssurvey}. As summarized in Table~\ref{tab:related_surveys}, our focus is different: we organize Gaussian Splatting as a reconstruction pipeline and study which task-specific costs limit real-time deployment.
}

\begin{table}[!t]
\centering
\footnotesize
\caption{\textcolor{red}{Positioning of this survey relative to representative Gaussian Splatting surveys.}}
\label{tab:related_surveys}
\setlength{\tabcolsep}{3pt}
\renewcommand{\arraystretch}{1.08}
\begin{tabular}{>{\raggedright\arraybackslash}p{0.20\linewidth}
                >{\centering\arraybackslash}p{0.08\linewidth}
                >{\raggedright\arraybackslash}p{0.31\linewidth}
                >{\raggedright\arraybackslash}p{0.32\linewidth}}
\toprule
\textbf{Survey} & \textbf{Year} & \textbf{Primary organization} & \textbf{Distinctive emphasis} \\
\midrule

Wu et al.~\cite{wu2024recent3dgs} &
2024 &
Reconstruction, editing, downstream applications &
Functional overview of 3DGS developments \\

Fei et al.~\cite{fei2025newera} &
2025 &
Characteristics and applications &
Theoretical foundations and application taxonomy \\

Bao et al.~\cite{bao2025gaussiansurvey} &
2025 &
Optimization, applications, extensions &
Technical modules, challenges, and opportunities \\

Bagdasarian et al.~\cite{bagdasarian2025gszip} &
2025 &
Compression and compaction &
Representation size and compression efficiency \\

Zhu et al.~\cite{zhu2024roboticssurvey} &
2024 &
Robotics applications and technical advances &
3DGS for robotic perception and interaction \\

\textbf{Ours} &
\textbf{2023--2025} &
\textbf{Five reconstruction-pipeline families} &
\textbf{Task-specific runtime, quantitative evaluation, and deployment trade-offs} \\
\bottomrule
\end{tabular}
\end{table}

{\color{red}
In this survey, \emph{real-time} is task specific. We distinguish render-only throughput from scene generation, dynamic or streaming updates, SLAM tracking and mapping, and offline optimization. These quantities are compared only when their unit of work and timing boundary are aligned, as defined in \Cref{sec:realtime}.
}

\begin{figure*}[t]
    \centering
    \includegraphics[width=\textwidth]{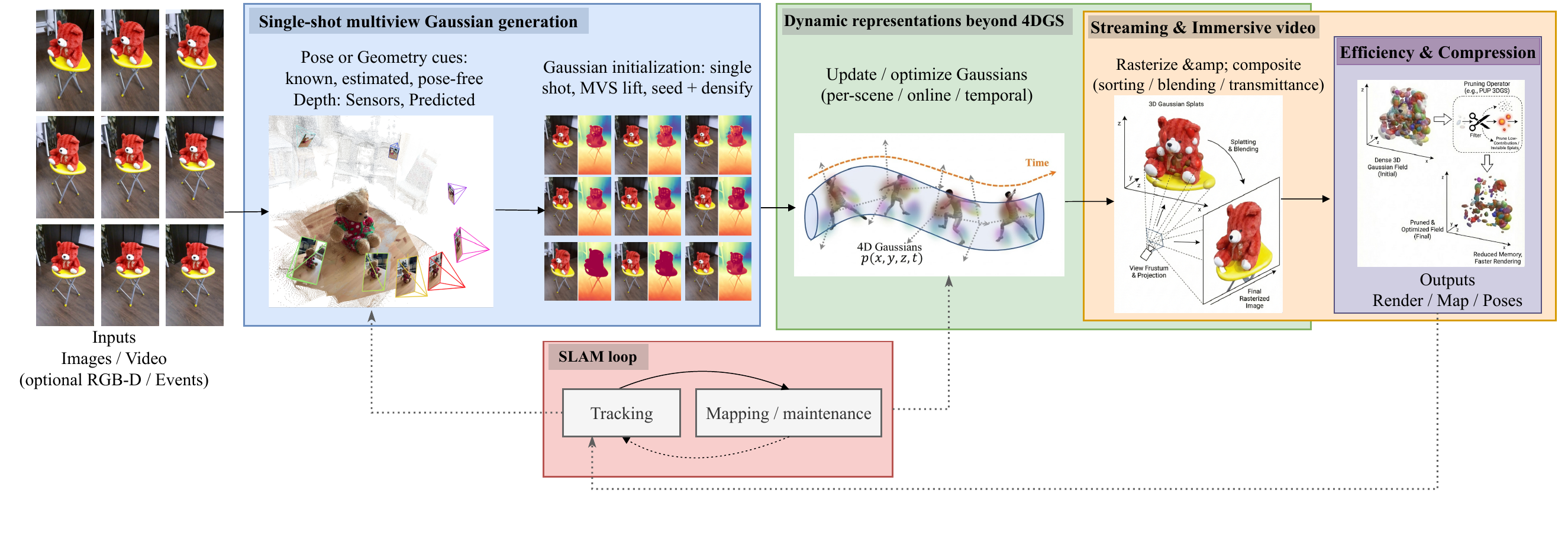}
    \vspace{-10 mm}
    \caption{Overview of a Gaussian Splatting–based reconstruction pipeline. Starting from images/video (optionally RGB-D/events), the system forms pose/geometry cues, initializes Gaussians (one-shot or MVS lifting), updates/optimizes the scene representation (per-scene, online, or temporal), and rasterizes/composites to produce renders, maps, and poses. Shaded containers indicate where the five taxonomy families primarily intervene (single-shot generation, dynamic representations, efficiency/compression, streaming/immersive video, and SLAM with tracking–mapping). }
    \label{fig:prelim}
\end{figure*}

{\color{red}
We formulate real-time Gaussian scene reconstruction from the pipeline perspective illustrated in \Cref{fig:prelim}. We organize the literature into five overlapping families according to the stage containing the dominant contribution: single-shot generation, dynamic representations, efficiency and compression, streaming and immersive systems, and Gaussian Splatting SLAM. This pipeline-centric view exposes where geometry, pose estimation, temporal updates, memory, and rendering costs arise across otherwise different GS systems.
}

We summarize the scope, organization, and contributions of this survey as follows:
\begin{itemize}

    {\color{red}
    \item \textbf{Pipeline-centric taxonomy:} We organize GS research into five overlapping families according to the dominant stage of the reconstruction pipeline, supported by a compact tag schema for structured comparison (\Cref{fig:taxonomy,tab:tags}).
 
    \item \textbf{Task-specific real-time framework:} We distinguish render-only throughput, scene generation, temporal updates, SLAM tracking and mapping, and offline optimization using explicit units of work and timing boundaries (\Cref{sec:realtime,sec:runtime_metrics,sec:eval_protocol}).

    \item \textbf{Quantitative and algorithmic evaluation:} We provide a controlled comparison of compatible sparse-view methods and complementary native evaluations across the remaining families, with explicit comparability boundaries and mechanism-level analysis (Section~6).
  
    \item \textbf{Cross-family synthesis:} We connect algorithmic choices to quality, latency, memory, update cost, and deployment suitability, and identify remaining barriers toward end-to-end real-time systems (Section~6 and \Cref{sec:open}).
    }

\end{itemize}
\section{Preliminaries} \label{sec:prelim}

\Cref{fig:prelim} summarizes the end-to-end Gaussian Splatting reconstruction pipeline and where our five taxonomy families intervene. This section fixes terminology and notation, summarizes the core 3DGS representation and rasterization/compositing used for real-time rendering, and defines the training regimes, input/supervision assumptions, and the real-time criterion used throughout. {\color{red}
We conclude with a compact tag schema used to structure the taxonomy and method comparisons.
}

\subsection{Scope and Terms}

We use \emph{Gaussian Splatting} to refer to modern \emph{3D Gaussian Splatting} (3DGS)~\cite{3dgs_og}: an explicit set of 3D Gaussian primitives parameterized in world space and rendered by fast (typically differentiable) screen-space splatting with rasterization/compositing. Under this scope, the primary object is a \emph{Gaussian scene representation} (centers, covariances, opacities, and appearance parameters), and the key efficiency advantage comes from rasterizing a finite set of primitives rather than volumetric ray marching.
We do \emph{not} survey NeRFs~\cite{mildenhall2021nerf} as a primary topic. We cite NeRF-related work only when necessary for context (e.g., datasets/metrics) or when a method explicitly transfers radiance-field ideas into Gaussian representations.

{\color{red}
Finally, we use \emph{real-time scene reconstruction} in a task-specific sense
rather than through a universal FPS threshold. Rendering, scene generation,
dynamic or streaming updates, and SLAM are evaluated using their corresponding
units of work and timing boundaries, as defined in \Cref{sec:realtime}.
}

\begin{table}[!b]
\centering
\footnotesize
\caption{Notation used throughout the survey.}
\label{tab:notation}
\setlength{\tabcolsep}{6pt}
\renewcommand{\arraystretch}{1.15}
\begin{tabularx}{\columnwidth}{@{}lX@{}}
\toprule
\textbf{Symbol} & \textbf{Meaning} \\
\midrule
$I_i$ / $I_t$ & Input image $i$ (or frame at time $t$). \\
$K$ & Camera intrinsics matrix. \\
$T_i=(R_i,t_i)$ & Camera pose (extrinsics) for view $i$; rotation $R_i$ and translation $t_i$. \\
$\pi(\cdot)$ & Camera projection function mapping 3D to image coordinates. \\
$\mathbf{u}$ & Pixel coordinate in the image plane. \\
$\mathcal{G}$ / $\mathcal{G}_t$ & Set of Gaussians for a static scene (or at time $t$). \\
$g_j$ & The $j$-th Gaussian primitive in $\mathcal{G}$. \\
$\mu_j \in \mathbb{R}^3$ & 3D mean (center) of Gaussian $g_j$ in world coordinates. \\
$\Sigma_j$ & 3D covariance of $g_j$ (often parameterized via scale + rotation). \\
$\alpha_j$ & Opacity (or density/weight) parameter of $g_j$. \\
$\mathbf{c}_j$ / $\mathbf{f}_j$ & Appearance parameters (e.g., color or feature coefficients such as SH). \\
$\hat{I}$ & Rendered image produced by rasterization/compositing. \\
$\hat{D}$ & Rendered depth (if reported/used by a method). \\
$N$ & Number of Gaussians in the scene representation. \\
\bottomrule
\end{tabularx}
\end{table}

\subsection{Notations}

\Cref{tab:notation} lists the main symbols used in the survey. For dynamic scenes, quantities are indexed by time $t$ and the Gaussian set at time $t$ is denoted $\mathcal{G}_t$. Unless stated otherwise, each primitive is represented by $(\mu_j,\Sigma_j,\alpha_j,\mathbf{c}_j)$, where $\mathbf{c}_j$ may be RGB, spherical harmonics coefficients, or learned features depending on the method.

\subsection{Gaussian Representation}
\label{sec:gaussian_representation}

Gaussian Splatting represents a scene as $\mathcal{G}=\{g_j\}_{j=1}^{N}$, a set of $N$ anisotropic 3D Gaussian primitives in world coordinates. Each $g_j$ is defined by a mean $\mu_j\in\mathbb{R}^3$, covariance $\Sigma_j\in\mathbb{R}^{3\times 3}$ (commonly parameterized via rotation and per-axis scales for stability), opacity $\alpha_j\in[0,1]$, and appearance parameters $\mathbf{c}_j$. This explicit primitive-based representation enables high-throughput rasterization pipelines compared to radiance-field rendering~\cite{3dgs_og}.

Methods primarily differ in (i) which parameters are optimized vs.\ predicted, and (ii) how time is modeled. Static pipelines often optimize all parameters per scene with SfM/MVS/depth-based initialization; dynamic extensions introduce time dependence via explicit per-frame parameters, deformation models, or static/dynamic partitions. Online/SLAM-style systems additionally constrain updates to bounded latency and control representation growth via pruning, keyframes, and local refinement.

\paragraph{Representation complexity and real-time knobs}
Runtime and memory are dominated by the number of primitives $N$ and the appearance bandwidth (e.g., SH degree or feature dimensionality). Dynamic/streaming settings add growth over time, so many systems explicitly regulate $N$ and appearance parameters to meet real-time targets.

\paragraph{Representation vs.\ serialization (file formats)}
Gaussian scenes may be stored in different encodings (e.g., PLY-like formats or \texttt{.splat}-style binaries). These typically change serialization/layout and attribute encoding (e.g., quantization, SH/features) rather than the underlying representation $\{(\mu_j,\Sigma_j,\alpha_j,\mathbf{c}_j)\}_{j=1}^{N}$.

\subsection{Rasterization}

Given camera intrinsics $K$ and pose $T=(R,t)$, each mean $\mu_j$ projects to
\begin{equation}
    \hat{\mu}_j = \pi\!\big(K, T; \mu_j\big) \in \mathbb{R}^2,
\end{equation}
and its 3D covariance $\Sigma_j$ induces an elliptical screen-space footprint (2D covariance) $\Sigma^{2D}_j$. Rendering composites projected splats in approximate front-to-back order using alpha/transmittance blending~\cite{3dgs_og}. Modern implementations use GPU tiled rasterization and sorting so only splats overlapping a pixel/tile are processed, which is central to real-time performance~\cite{ye2025gsplat}. Approximate visibility/sorting can cause artifacts in dense or dynamic regions~\cite{kheradmand2025stochasticsplats}, motivating improved compositing and hardware/system optimizations~\cite{yuan2025differentiablehardware3dgs,talegaonkar2025volumetrically,hanson2025speedysplat}.

\subsection{Training}

We group common regimes by \emph{when} optimization/learning happens: \textbf{per-scene/per-sequence optimization} (offline fitting), \textbf{amortized feed-forward generation} via pretrained models, and \textbf{dynamic/online} settings where temporal modeling and incremental updates impose bounded-latency constraints. Per-scene optimization fits $\{(\mu_j,\Sigma_j,\alpha_j,\mathbf{c}_j)\}$ for a specific sequence~\cite{3dgs_og}; amortized methods predict Gaussians at inference time (optionally followed by brief test-time refinement)~\cite{Xu2024DepthSplatCG,Chen2024MVSplatE3,Charatan2023PixelSplat3G,lin2025depthanything3recoveringvisual}. Dynamic methods often introduce time-dependent parameterizations or deformations and are frequently optimized per sequence~\cite{4dgs_og,e4dgs,Yang2023Deformable3G}, while online/SLAM pipelines prioritize training-free or pretrained components at runtime with efficient map maintenance.

\subsection{Inputs and Supervision}

We categorize input modalities as \textbf{multi-view} captures, \textbf{monocular video}, and \textbf{RGB-D} or sensor-augmented streams (e.g., events). Pose assumptions range from \textbf{known} (calibration/SfM) to \textbf{estimated/refined} (tracked or jointly optimized) to \textbf{pose-free}. Supervision is often self-supervised via photometric consistency, but many pipelines incorporate pretrained priors (especially depth/MVS) to stabilize geometry and accelerate initialization.

{\color{red}
\subsection{Task-Specific Real-Time Definition}
\label{sec:realtime}

The term \emph{real-time} is used differently across Gaussian Splatting methods. Rendering, scene generation, streaming updates, and SLAM correspond to different tasks and should therefore not share a single throughput definition. We use approximately 24 FPS only as a reference for interactive playback, not as a universal threshold for reconstruction or scene updates.

Let all latency values be expressed in milliseconds. Render-only throughput is defined as

\begin{equation}
\mathrm{rFPS}
=
\frac{1000}{t_{\mathrm{render}}},
\end{equation}

where $t_{\mathrm{render}}$ is the time required to render one frame from an existing Gaussian representation. For a reconstruction or update task, the pipeline latency can be written as

\begin{equation}
t_{\mathrm{pipe}}
=
t_{\mathrm{pre}}
+
t_{\mathrm{pose}}
+
t_{\mathrm{depth/fusion}}
+
t_{\mathrm{init}}
+
t_{\mathrm{update}}
+
t_{\mathrm{render}},
\end{equation}

where only stages relevant to the evaluated task are included. The corresponding task throughput is

\begin{equation}
R_{\mathcal{T}}
=
\frac{1000}{t_{\mathcal{T}}}.
\end{equation}

Accordingly, we report scene throughput for feed-forward reconstruction, update throughput for dynamic and streaming methods, and tracking or mapping throughput for SLAM systems. Offline optimization is reported using training or optimization time rather than scenes per second. These quantities are compared only when the task, unit of work, and timing boundary are aligned.
}
\subsection{Summary of Tags}

To compare methods consistently across our taxonomy, we annotate each paper using a compact tag set capturing input assumptions, regime, temporal scope, and runtime setting. \Cref{tab:tags} summarizes the master schema and family-specific extensions.
\begin{table}[!t]
\centering
\footnotesize
\caption{Master tag schema used across the survey. Core tags are shared across all method families; family-specific tags are used only in the corresponding matrices.}
\label{tab:tags}
\setlength{\tabcolsep}{4pt}
\renewcommand{\arraystretch}{1.1}
\begin{tabularx}{\columnwidth}{@{}l X@{}}
\toprule
\multicolumn{2}{@{}l}{\textbf{Core tags (shared vocabulary)}} \\ \midrule
Input            & mono-video; sparse-mv; dense-mv; RGB-D; events+RGB \\
Pose             & known; estimated/refined; pose-free \\
Depth cue        & sensor depth; predicted depth; MVS; none \\
Regime           & PerSeq (per-seq. opt.); PreTr (amortized); HybridR (init+refine) \\
Temporal scope   & static; dynamic \\
Runtime setting  & offline; online; streaming \\
Primary output   & rendering/NVS; mapping; poses; immersive video \\
\midrule
\multicolumn{2}{@{}l}{\textbf{Dynamic-family tags }} \\ \midrule
Type             & Def.; KeyT; Struct.; 4DPrim; Hybrid; FreeT \\
Dynamics         & non-rigid; articulated; multi-object; occl./disoccl. \\
Online           & Yes / No \\
RT render        & Yes / No / TBA \\
\midrule
\multicolumn{2}{@{}l}{\textbf{Single-shot-family tags }} \\ \midrule
Input / Pose / Depth / Regime 
                 & Uses the core tags above (exact column match) \\
\midrule
\multicolumn{2}{@{}l}{\textbf{Efficiency-family tags }} \\ \midrule
Lever            & Prune; Quant; LoD; Rend; HW; Sys \\
Scope            & Static; Dynamic; Both \\
Offline preproc. & Yes / No \\
RT               & Yes / No \\
\bottomrule
\end{tabularx}
\end{table}

\section{Methodology}
\label{sec:method}

{\color{red}
We survey Gaussian Splatting (GS) research toward real-time scene reconstruction using the task-specific definitions in \Cref{sec:realtime}.
}
Our workflow is: (i) collect papers broadly, (ii) apply inclusion criteria aligned with the GS reconstruction pipeline, (iii) assign a pipeline-centric taxonomy with consistent tags, and (iv) interpret reported efficiency using end-to-end-aware conventions.

\subsection{Literature Collection and Inclusion Criteria}

We prioritize recent work (2023--2025) while retaining foundational baselines needed for context (datasets, metrics, and core dynamic/NVS references).
{\color{red}
Literature was identified primarily through Google Scholar and the proceedings of CVPR, ICCV, ECCV, NeurIPS, and ICLR.
Seed queries included \emph{Gaussian splatting}, \emph{Gaussian splatting for SLAM}, \emph{efficient Gaussian splatting}, \emph{fast Gaussian splatting}, \emph{sparse-view Gaussian splatting}, and \emph{zero-shot Gaussian splatting}, together with closely related variants for dynamic reconstruction, streaming, compression, pose-free reconstruction, and real-time reconstruction.
Reference lists of relevant papers were also inspected to identify related and foundational work (backward citation snowballing).
}
We include methods whose \emph{primary scene representation} is an explicit Gaussian-based model (or a direct extension) and that materially affect one or more stages of a GS reconstruction pipeline (initialization/generation, temporal modeling, map maintenance, streaming/delivery, compositing, or tracking-and-mapping).
We exclude NeRF-style methods as a primary focus, citing them only when necessary for context or when they explicitly bridge into Gaussian representations.
{\color{red}
Duplicate versions and works outside the surveyed reconstruction/system scope are excluded; dataset, metric, and foundational papers are retained as supporting references rather than surveyed GS methods.
Because the literature search evolved iteratively during preparation of the survey, intermediate retrieval and screening counts were not recorded prospectively and are therefore not reconstructed retrospectively.
}

\subsection{Tagging and Taxonomy Assignment}

\begin{figure*}[!t]
    \centering
    \includegraphics[width=\textwidth]{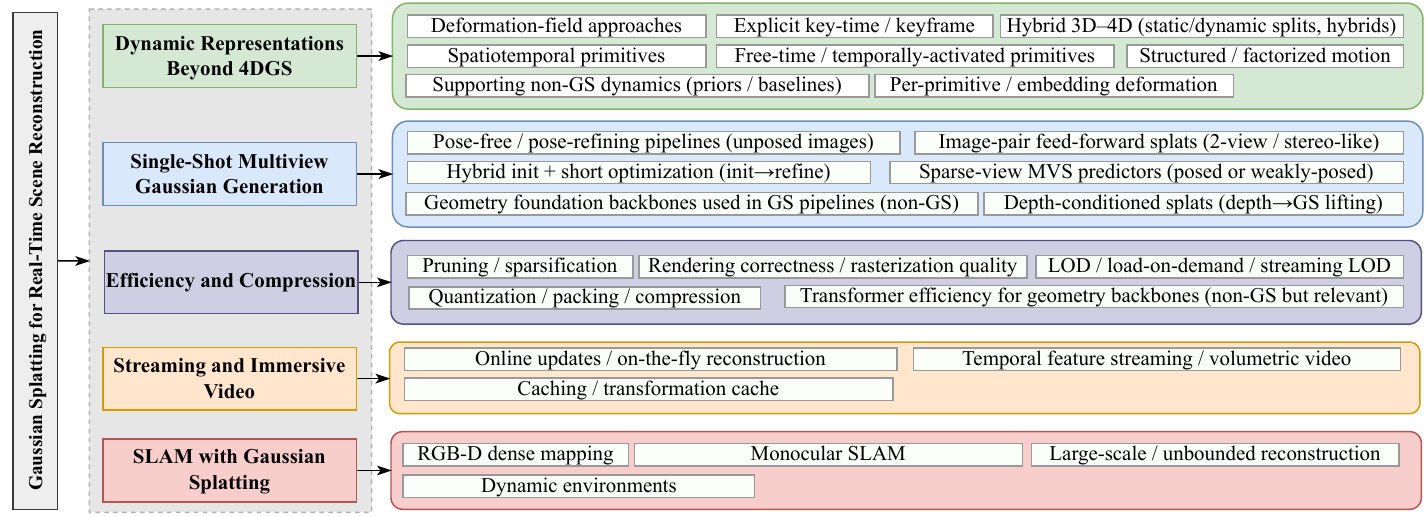}
    \vspace{-5 mm}
\caption{Taxonomy of Gaussian Splatting for Real-Time Scene Reconstruction.
The diagram categorizes methods into dynamic representations, single-shot generation,
efficiency, \textcolor{red}{streaming and immersive video,} and SLAM.
\textcolor{red}{For a clearer chronological view and additional representative methods
from each family, see \Cref{fig:timeline}.}}
\label{fig:taxonomy}
\end{figure*}

Each paper is annotated using the master tag schema in Table~\ref{tab:tags} (input/pose/depth cues, regime, temporal scope, runtime setting, and efficiency levers).
We assign each method to one \emph{primary} family based on its dominant contribution, while noting meaningful overlaps when a method intervenes in multiple pipeline stages.
The five families in Fig.~\ref{fig:taxonomy} are constructed to reflect \emph{where} methods intervene operationally in the end-to-end pipeline (Fig.~\ref{fig:prelim}), enabling consistent comparison across tables and sections.

\subsection{Interpreting Performance Claims}

{\color{red}
To avoid conflating rendering speed with reconstruction throughput, runtime evidence is interpreted according to the task being measured rather than through a universal FPS or SPS quantity.
Render-only results, scene generation, dynamic or streaming updates, SLAM, and offline optimization follow the timing boundaries defined in \Cref{sec:realtime}.
Results are treated as directly comparable only when the dataset, input protocol, resolution, hardware, unit of work, and timing boundary are sufficiently aligned; otherwise, they are treated as method-specific evidence.
}
When code, pretrained models, or timing scripts are available, we prioritize these artifacts over high-level paper claims for interpreting runtime and memory.
{\color{red}
For methods evaluated in Section~6, repository configurations, checkpoints, native schedules, and evaluation procedures are used where available to establish the reported operating conditions.
}

\section{Datasets and Evaluation Protocols}
\label{sec:reporting}

{\color{red}
This section summarizes the datasets, quality metrics, and evaluation protocols used throughout the survey. 
Runtime comparisons follow the task-specific definitions introduced in Section~\ref{sec:realtime}.
}

\subsection{Datasets and Quality Metrics}
\label{sec:datasets_metrics}

\paragraph{Datasets (representative regimes)}
Although datasets are not the main focus of this survey, they strongly influence what claims are comparable.
Across the literature, benchmarks cluster into five recurring regimes.
\textbf{(i) Multi-view dome/studio captures} provide calibrated multi-camera observations and controlled dynamics,
commonly used for high-fidelity dynamic reconstruction (e.g., Technicolor Light Field \cite{technicolor},
Neu3DV \cite{neu3dv}, Replay \cite{shapovalov2023replay}, NeRSemble \cite{kirschstein2023nersemble},
PKU-DyMVHumans \cite{zheng2024pku}, DiVa-360 \cite{lu2024diva}).
\textbf{(ii) Monocular/handheld sequences} emphasize pose uncertainty, motion blur, and in-the-wild dynamics,
often inherited from dynamic NeRF benchmarks (e.g., NSFF Dynamic \cite{nsff-dynamic}, D-NeRF \cite{dnerf},
Nerfies \cite{park2021nerfies}, HyperNeRF \cite{Park2021HyperNeRF}) and more recent GS/video-generation settings
(e.g., Splat4D \cite{yin2025splat4d}).
\textbf{(iii) Event+RGB settings} target fast motion and low-light where conventional frames degrade, using event
streams as additional supervision or cues (e.g., Dynamic EventNeRF \cite{Rudnev2024DynamicER}, E-4DGS \cite{e4dgs}).
\textbf{(iv) Outdoor/driving/drone captures} evaluate scalability and SLAM-like conditions, often with LiDAR/SLAM pose
supervision and dynamic agents (e.g., MTGS \cite{li2025mtgs}, DroneSplat \cite{Tang2025DroneSplat3G},
Wide-Range-4D \cite{yang2025widerange4d}).
\textbf{(v) Synthetic/physics-based datasets} provide controllable dynamics and/or ground-truth structure for
stress-testing temporal modeling (e.g., Deep3D Mask Volume \cite{d3dmask}, PhysGaia \cite{kim2025physgaia},
Hybrid 3D-4DGS \cite{oh2025hybrid}, DynaSplat \cite{deng2025dynasplat}).

\paragraph{Image fidelity and temporal stability}
Most works report standard per-frame fidelity using PSNR, SSIM \cite{wang2004ssim}, and LPIPS \cite{zhang2018lpips},
typically averaged over time for dynamic sequences. To reflect \emph{temporal stability}, several papers additionally
report flow-warped temporal errors or flow-based temporal discrepancy scores \cite{chu2020temporal}.
Some works also use video metrics such as FVD \cite{unterthiner2018fvd}, although these are less consistently reported.

\subsection{Task-Specific Runtime Metrics}
\label{sec:runtime_metrics}

{\color{red}
Runtime is reported according to the task being measured. Render FPS (rFPS) refers only to rendering an existing Gaussian representation. Feed-forward methods use scene throughput, dynamic and streaming methods use update throughput, and SLAM methods use tracking or mapping throughput. Offline optimization is reported using training or optimization time. These quantities are compared only when they measure the same task under the same timing boundary.
}

{\color{red}
\begin{table}[t]
\centering
\caption{\textcolor{red}{Task-specific runtime quantities used in this survey.}}
\label{tab:task_runtime}
\small
\begin{tabular}{p{2.6cm} p{2.5cm} p{2.8cm} p{2.3cm}}
\hline
\textbf{Task} & \textbf{Unit of work} & \textbf{Timing boundary} & \textbf{Metric} \\
\hline
Rendering &
One rendered frame &
Existing representation to rendered frame &
rFPS \\

Scene generation &
One fixed input-view set &
Input views to Gaussian representation &
Scene throughput \\

Dynamic / streaming &
One defined update or transition &
Update input to updated representation &
Update throughput \\

SLAM tracking &
One input frame &
Frame input to pose estimate &
Tracking FPS \\

SLAM mapping &
One map update &
Mapping input to updated map &
Mapping throughput \\

Offline optimization &
One training or optimization run &
Start to completed model/checkpoint &
Time / iterations \\
\hline
\end{tabular}
\end{table}
}

\subsection{Evaluation Protocol}
\label{sec:eval_protocol}

{\color{red}
Quantitative comparisons require aligned datasets, input settings, resolution, hardware, and timing boundaries. When these conditions are not aligned, results are treated as method-specific evidence rather than direct comparisons. We therefore report both quality and runtime together with the experimental conditions under which they were measured.
}
\section{Taxonomy}
\label{sec:taxonomy}

\begin{figure*}[!t]
    \centering
    \includegraphics[width=\textwidth]{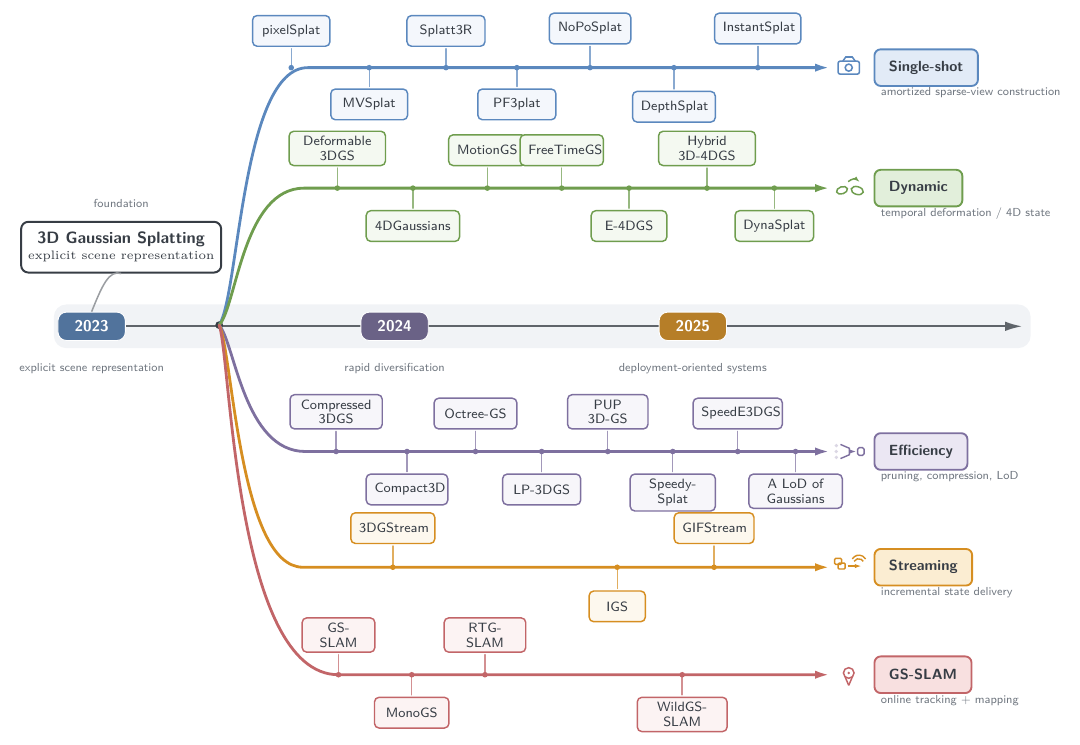}
    \caption{\textcolor{red}{Chronological development of representative Gaussian Splatting methods across the five survey families. Methods are positioned by publication year and primary contribution; the figure is representative rather than exhaustive.}}
    \label{fig:timeline}
\end{figure*}

We organize Gaussian Splatting work for real-time reconstruction into five pipeline-centric families, defined by the
primary stage a method intervenes in (capture/training, representation, rendering, system delivery, or online estimation).
Because many methods are multi-faceted, we allow secondary overlaps but assign each paper a single \emph{primary} family
based on its dominant technical contribution.

{\color{red}
Figure~\ref{fig:timeline} shows the chronological development of representative methods across these families, illustrating how work following 3DGS progressively expanded toward amortized generation, temporal modeling, efficiency, streaming, and online mapping.
}

\label{sec:taxonomy_families}

\paragraph{\textbf{Single-shot / amortized Gaussian generation}}
Feed-forward construction of a usable Gaussian scene from one or few images, often using depth/MVS priors or learned
regularizers, optionally followed by brief refinement. Poses may be known, refined, or pose-free, but the defining feature
is replacing heavy per-scene iterative fitting with amortized inference from limited views.

\paragraph{\textbf{Dynamic representations beyond 4DGS}}
Methods that treat time as a first-class variable in the scene model (e.g., time-conditioned parameters, deformation/warp
fields, key-time primitives, canonical-plus-delta factorization) or explicitly mitigate dynamic failure modes such as
ghosting, identity swaps, occlusion/disocclusion artifacts, and temporal flicker. The emphasis is \emph{modeling} temporal
change, not merely delivering or replaying content over time.

\paragraph{\textbf{Efficiency and compression}}
Works whose central goal is reducing compute, memory, or storage/bandwidth while preserving real-time operation, via levers
such as pruning/sparsification, quantization/packing, view-adaptive LoD selection, or system/hardware acceleration. We place
a method here when efficiency is the primary objective rather than a side effect.

\paragraph{\textbf{Streaming and immersive video systems}}
Long-duration capture and interactive free-viewpoint playback under latency/bandwidth constraints, enabled by explicit
delivery mechanisms (progressive transmission, caching/tiling/segmenting, rate control, bandwidth-aware loading, or
continual update protocols). The core contribution is \emph{system delivery and interactivity over time}, not new temporal
scene parameterizations.

\paragraph{\textbf{SLAM with Gaussian Splatting}}
Online tracking-and-mapping where camera poses are estimated incrementally and the Gaussian map is updated under bounded
latency/compute, typically using keyframes, map maintenance/pruning, robustness to blur/occlusions, and (when relevant)
loop-consistency considerations. We place a method here when meeting tracking deadlines and maintaining a bounded-cost map
are central to the design.
\section{Quantitative Evaluation and Synthesis}
\subsection{Single-Shot and Sparse-View Generation}

\begin{table}[t]
\centering
\caption{\textcolor{red}{Controlled comparison of single-shot Gaussian reconstruction methods on
RealEstate10K~\cite{zhou2018stereo}. All methods use the same 100 test scenes,
two context views, three target views, $256\times256$ resolution, and an
NVIDIA A100 GPU. Model-stage scene latency and scene throughput include the
encoder and three target decodes, while rFPS reports decoder-only rendering.}}
\label{tab:single_shot_controlled}

\small
\setlength{\tabcolsep}{2.5pt}
\renewcommand{\arraystretch}{1.08}

\begin{tabular}{
>{\raggedright\arraybackslash}p{1.95cm}
c
c
c
>{\centering\arraybackslash}p{2cm}
c
c
>{\centering\arraybackslash}p{0.95cm}
}

\hline
\textbf{Method}
& \textbf{PSNR $\uparrow$}
& \textbf{SSIM $\uparrow$}
& \textbf{LPIPS $\downarrow$}
& \textbf{Model-stage latency (ms) $\downarrow$}
\textbf{Model-stage scene/s $\uparrow$}
& \textbf{rFPS $\uparrow$}
& \textbf{GPU (GiB) $\downarrow$} \\
\hline

MVSplat~\cite{Chen2024MVSplatE3}
& 26.78
& 0.881
& 0.119
& \textbf{37.63}
& \textbf{26.57}
& \textbf{871.21}
& \textbf{1.18} \\

DepthSplat~\cite{Xu2024DepthSplatCG}
& \textbf{27.62}
& \textbf{0.898}
& \textbf{0.106}
& 65.37
& 15.30
& 654.68
& 2.63 \\

pixelSplat~\cite{Charatan2023PixelSplat3G}
& 26.40
& 0.873
& 0.127
& 120.57
& 8.29
& 462.29
& 2.77 \\

\hline
\end{tabular}
\end{table}

{\color{red}
\begin{table}[t]
\centering
\caption{\textcolor{red}{Family-specific measurements for additional single-shot and sparse-view methods.
These results use different datasets and execution settings than
\cref{tab:single_shot_controlled} and are not directly comparable.
Runtime refers to the method-specific measured construction or optimization stages.}}

\label{tab:single_shot_native}


\small
\setlength{\tabcolsep}{2pt}
\renewcommand{\arraystretch}{1.08}
\begin{tabular}{
>{\raggedright\arraybackslash}p{1.99cm}
>{\raggedright\arraybackslash}p{2.85cm}
c
c
c
>{\raggedright\arraybackslash}p{2.65cm}
}
\hline
\textbf{Method}
& \textbf{Dataset / input}
& \textbf{PSNR $\uparrow$}
& \textbf{SSIM $\uparrow$}
& \textbf{LPIPS $\downarrow$}
& \textbf{Runtime} \\
\hline

Splatt3R~\cite{smart2024splatt3r}
& ScanNet++~\cite{yeshwanth2023scannetpp}, 2 views
& --
& --
& --
& 1.10 s construction / export \\

InstantSplat~\cite{fan2025instantsplatsparseviewgaussiansplatting}
& T\&T Horse~\cite{knapitsch2017tanks}, 3 views
& 23.30
& 0.823
& 0.125
& 16.01 s initialization + 35.02 s optimization \\

\hline
\end{tabular}
\end{table}
}
{\color{red}
For calibrated sparse-view methods, Gaussian construction can be expressed as
depth inference followed by unprojection,
\begin{equation}
\mu_i(\mathbf{u})
=
\pi^{-1}\!\left(
\mathbf{u},\hat{d}_i(\mathbf{u});K_i,T_i
\right),
\qquad
(\Sigma_i,\alpha_i,\mathbf{c}_i)
=
H_{\theta}\!\left(\mathbf{f}_i(\mathbf{u})\right).
\end{equation}
MVSplat~\cite{Chen2024MVSplatE3} estimates $\hat d_i$ from a plane-sweep cost
volume,
\begin{equation}
C_i(\mathbf{u},d_k)
=
\operatorname{Agg}_{m\neq i}
\left\langle
F_i(\mathbf{u}),
\mathcal{W}_{m\rightarrow i}(F_m;d_k)
\right\rangle ,
\qquad
\hat d_i
=
\sum_k
\operatorname{softmax}(C_i)_k d_k ,
\end{equation}
so geometry is determined primarily by cross-view correspondence.
DepthSplat~\cite{Xu2024DepthSplatCG} instead augments the multi-view matching
representation with pretrained monocular depth features,
\begin{equation}
\widetilde C_i
=
\Phi_{\theta}
\!\left(
C_i,F_i^{\mathrm{mono}}
\right),
\end{equation}
before depth regression and Gaussian unprojection. This additional geometric
prior corresponds to the highest PSNR/SSIM and lowest LPIPS in
\cref{tab:single_shot_controlled}, but also increases model-stage latency and
memory relative to MVSplat. pixelSplat~\cite{Charatan2023PixelSplat3G} avoids
committing directly to one depth by predicting
$p_i(d\mid\mathbf{u})$ and differentiably sampling
\begin{equation}
d_i \sim p_i(d\mid\mathbf{u}),
\qquad
\mu_i(\mathbf{u})
=
\pi^{-1}(\mathbf{u},d_i;K_i,T_i).
\end{equation}
Its more flexible treatment of geometric ambiguity is accompanied by the
largest construction latency and memory footprint in the controlled experiment.
Thus, \cref{tab:single_shot_controlled} shows that the mechanism used to
estimate $\mu_j$ directly affects both reconstruction quality and construction
cost; in this experiment, augmenting multi-view matching with monocular depth
features gives the strongest quality, while the lighter cost-volume formulation
gives the lowest latency.
}

{\color{red}
The native methods in \cref{tab:single_shot_native} remove the calibrated-pose
assumption in different ways. Splatt3R~\cite{smart2024splatt3r} extends MASt3R \cite{leroy2024grounding}
with a Gaussian prediction head, which can be summarized as
\begin{equation}
\left(
\mathbf{x}_j,\,
\Delta\mathbf{x}_j,\,
q_j,\,
s_j,\,
\alpha_j,\,
\mathbf{c}_j
\right)
=
H_{\theta}(I_1,I_2),
\qquad
\mu_j=\mathbf{x}_j+\Delta\mathbf{x}_j ,
\end{equation}
with $\Sigma_j$ obtained from the predicted rotation $q_j$ and scale $s_j$.
The scene is therefore produced directly from an uncalibrated pair without
optimizing camera variables. InstantSplat~
\cite{fan2025instantsplatsparseviewgaussiansplatting} instead uses dense stereo
priors to initialize both geometry and cameras and then performs Gaussian Bundle
Adjustment,
\begin{equation}
\min_{\mathcal{G},\,\{T_i\}}
\sum_i
\mathcal{L}_{\mathrm{photo}}
\left(
I_i,\hat I(\mathcal{G},T_i)
\right),
\end{equation}
jointly updating the Gaussian parameters and poses. This distinction is visible
quantitatively: Splatt3R requires 1.10\,s for measured construction/export,
whereas InstantSplat uses 16.01\,s for initialization followed by 35.02\,s of
optimization. Hence the former admits a construction throughput, while the
latter is an optimization problem whose runtime depends on the refinement
budget. In both cases, the much larger decoder-only rFPS reported in
\cref{tab:single_shot_controlled} remains a separate quantity from the cost of
constructing $\mathcal{G}$.
Recent feed-forward generalists broaden this regime to unconstrained input views
and unified geometry-to-Gaussian prediction, as exemplified by AnySplat and
YoNoSplat~\cite{Jiang2025AnySplatF3,ye2025yonosplat}. These systems relax capture
assumptions, but move additional latency and memory pressure into their learned
geometry backbones and cross-view fusion.
}
\subsection{Dynamic Representations}

{\color{red}
Let $q_j$ and $s_j$ denote the rotation and scale parameterization of
$\Sigma_j$. Dynamic deformation can then be written as
\begin{equation}
(\Delta\mu_j,\Delta q_j,\Delta s_j)
=
F_{\theta}(\mu_j,t),
\qquad
\mu_j(t)=\mu_j+\Delta\mu_j,
\end{equation}
with the predicted rotation and scale determining $\Sigma_j(t)$. Deformable-3D-Gaussians~\cite{Yang2023Deformable3G}
uses an MLP deformation field conditioned directly on canonical position and time.
4DGaussians~\cite{4dgs_og} instead obtains the deformation from factorized
spatio-temporal features, using a HexPlane-style ~\cite{Cao2023HexPlaneAF} 4D encoding followed by a
lightweight decoder. Thus, the former places motion complexity in the continuous
deformation network, whereas the latter shifts part of it into an explicit
space--time feature structure. In \cref{tab:dynamic_native}, both reach similar
quality on D-NeRF \textit{bouncingballs}: 41.38\,dB for Deformable-3D-Gaussians
and 40.81\,dB for 4DGaussians. These values show that both deformation
parameterizations can represent the same synthetic motion regime at high fidelity.
4DGaussians additionally renders at 36.03 rFPS in the measured native setup,
showing that the deformation representation can remain compatible with interactive
rendering after optimization.
}

\begin{table}[t]
\centering
\caption{\textcolor{red}{Native evaluation of dynamic Gaussian representation methods.
4DGaussians and Deformable-3D-Gaussians are evaluated on D-NeRF
\textit{bouncingballs}~\cite{dnerf}; MotionGS was evaluated on the DyNeRF
\textit{flame\_steak} sequence~\cite{neu3dv} using its documented schedule but did
not complete. Results use method-specific native evaluation procedures and are not
a controlled cross-method benchmark.}}
\label{tab:dynamic_native}

\small
\setlength{\tabcolsep}{2.5pt}
\renewcommand{\arraystretch}{1.12}

\begin{tabular}{
p{2.15cm}
p{1.95cm}
p{1.2cm}
c
c
p{1.55cm}
p{0.95cm}
}
\hline
\textbf{Method}
& \textbf{Dataset}
& \textbf{Schedule}
& \textbf{PSNR $\uparrow$}
& \textbf{SSIM $\uparrow$}
& \textbf{LPIPS $\downarrow$}
& \textbf{rFPS $\uparrow$} \\
\hline

4DGaussians~\cite{4dgs_og}
& D-NeRF~\cite{dnerf}
& 20k
& 40.81
& 0.994
& 0.014 VGG / 0.006 Alex
& 36.03 \\

Deformable-3D-Gaussians~\cite{Yang2023Deformable3G}
& D-NeRF~\cite{dnerf}
& 40k
& 41.38
& 0.996
& 0.008
& -- \\

MotionGS~\cite{zhu2024motiongs}
& DyNeRF~\cite{neu3dv}
& 20k
& --
& --
& --
& -- \\

\hline
\end{tabular}

\vspace{1mm}

\footnotesize
MotionGS exhausted the 80\,GB GPU memory at iteration 7,600 and therefore
produced no completed checkpoint or final quality result.
\end{table}

{\color{red}
MotionGS~\cite{zhu2024motiongs} modifies this formulation by adding explicit motion
supervision to the deformation field. Optical flow is decomposed as
\begin{equation}
F^{M}_{t\rightarrow t+1}
=
F_{t\rightarrow t+1}
-
F^{C}_{t\rightarrow t+1},
\end{equation}
where $F^{C}$ is the component induced by camera motion and $F^{M}$ isolates
object motion. The predicted 3D Gaussian deformation is projected into Gaussian
flow $F^{G}$ and constrained by
\begin{equation}
\mathcal{L}_{\mathrm{flow}}
=
\left\|
\operatorname{sg}(F^{M}_{t\rightarrow t+1})
-
F^{G}_{t\rightarrow t+1}
\right\|_{1}.
\end{equation}
This changes the optimization problem from learning motion primarily through
photometric reconstruction to explicitly matching projected Gaussian motion.
This supervision couples deformation accuracy to rendered depth and camera
pose: errors in either term perturb $F^{C}$ and therefore $F^{M}$. In our
native DyNeRF \textit{flame\_steak} run, the documented 20k schedule exhausted
an 80\,GB A100 at iteration 7,600 (\cref{tab:dynamic_native}), showing that the
complete motion-guided pipeline can become memory-bound before convergence.
Other important dynamic formulations place temporal structure more explicitly:
4D-RotorGS and Ex4DGS encode time directly in the primitive representation,
whereas SC-GS uses sparse control points to structure deformation and support
motion editing~\cite{rotorgs,ex4dgs,scgs}. These alternatives reduce reliance on
one unconstrained per-Gaussian warp, but introduce their own temporal-parameter
or control-structure costs.
}


\subsection{Efficiency and Compression}

\begin{table}[t]
\centering
\caption{\textcolor{red}{Native evaluation of efficiency and compression methods.
Results use method-specific training schedules. Methods evaluated on different datasets
or native pipelines are not directly ranked against each other. rFPS is omitted when it is not reported by the native evaluation.}}
\label{tab:efficiency_native}

\small
\setlength{\tabcolsep}{2.2pt}
\renewcommand{\arraystretch}{1.10}

\begin{tabular}{
>{\raggedright\arraybackslash}p{2.5cm}
>{\raggedright\arraybackslash}p{1.75cm}
>{\raggedright\arraybackslash}p{1.75cm}
>{\centering\arraybackslash}p{1.1cm}
>{\centering\arraybackslash}p{1.1cm}
>{\centering\arraybackslash}p{1.2cm}
>{\centering\arraybackslash}p{0.95cm}
}
\hline
\textbf{Method}
& \textbf{Dataset}
& \textbf{Schedule}
& \textbf{PSNR $\uparrow$}
& \textbf{SSIM $\uparrow$}
& \textbf{LPIPS $\downarrow$}
& \textbf{rFPS $\uparrow$} \\
\hline

Octree-GS~\cite{streaminglod}
& N3DV~\cite{neu3dv}
& 40k
& 30.42
& 0.941
& 0.109
& -- \\

3D-GS~\cite{3dgs_og}
& N3DV~\cite{neu3dv}
& 30k
& 25.01
& 0.933
& 0.165
& 193.46 \\

PUP 3D-GS~\cite{HansonTuPUP3DGS}
& N3DV~\cite{neu3dv}
& 30k + 5k + 5k
& 25.42
& 0.930
& 0.185
& 155.15 \\

Compact3D~\cite{navaneet2024compact3d}
& N3DV~\cite{neu3dv}
& 30k base + 5k fine-tune
& 25.08
& 0.916
& 0.166
& -- \\

gsplat~\cite{ye2025gsplat}
& N3DV~\cite{neu3dv}
& 30k
& 21.93
& 0.919
& 0.208
& 119.31 \\

SpeedE3DGS~\cite{TuYing2025speede3dgs}
& D-NeRF~\cite{dnerf}
& 30k
& 38.59
& 0.993
& 0.017
& 762.25 \\

\hline
\end{tabular}
\end{table}

\begin{table}[t]
\centering
\caption{\textcolor{red}{Compression results from the shared 30k vanilla 3D-GS baseline
on N3DV \textit{flame\_steak}~\cite{neu3dv}. PUP 3D-GS and Compact3D use the same
trained baseline, allowing a direct comparison of representation size and retained quality.}}
\label{tab:compression_shared}

\small
\setlength{\tabcolsep}{4pt}
\renewcommand{\arraystretch}{1.08}

\begin{tabular}{lcccc}
\hline
\textbf{Representation}
& \textbf{Size (MiB) $\downarrow$}
& \textbf{Gaussians}
& \textbf{Size reduction $\uparrow$}
& \textbf{PSNR $\uparrow$} \\
\hline

3D-GS~\cite{3dgs_og}
& 75.50
& 319,222
& --
& 25.01 \\

PUP 3D-GS~\cite{HansonTuPUP3DGS}
& 7.55
& 31,922
& 90.00\%
& 25.42 \\

Compact3D~\cite{navaneet2024compact3d}
& 5.54
& --
& 92.66\%
& 25.08 \\

\hline
\end{tabular}
\end{table}
{\color{red}
For a Gaussian scene
$\mathcal{G}=\{g_j\}_{j=1}^{N}$, efficiency is governed by more than the primitive
count $N$. A useful conceptual decomposition is
\begin{equation}
B(\mathcal{G}) \approx N b_g,
\qquad
t_{\mathrm{render}}
\propto
c_g N_{\mathrm{vis}} + c_o M,
\end{equation}
where $b_g$ is the storage per Gaussian, $N_{\mathrm{vis}}$ is the number of
Gaussians processed for the current view, and $M$ denotes their accumulated
screen-space overlap. The methods in \cref{tab:efficiency_native,tab:compression_shared}
act on different terms of this model: pruning reduces $N$, quantization reduces
$b_g$, LoD reduces $N_{\mathrm{vis}}$, and renderer-level optimization reduces
the computational constants $c_g$ and $c_o$. Their reported ``efficiency'' gains
are therefore not interchangeable.
}

{\color{red}
PUP 3D-GS~\cite{HansonTuPUP3DGS} reduces $N$ using a second-order sensitivity
criterion. For spatial Gaussian parameters $\theta_j$, removing a primitive can
be interpreted through the local reconstruction-loss change
\begin{equation}
\Delta\mathcal{L}_j
\approx
\frac{1}{2}
\delta\theta_j^{\top} H_j \delta\theta_j ,
\end{equation}
where $H_j$ is the local Hessian block associated with the spatial
parameters of Gaussian $j$, 
so low-sensitivity Gaussians can be removed before refinement. On the shared
baseline in \cref{tab:compression_shared}, this removes approximately 90\% of
the Gaussians and 90\% of serialized bytes while PSNR changes only from
25.01 to 25.42\,dB. Compact3D~\cite{navaneet2024compact3d} instead primarily reduces
$b_g$ through vector quantization. If Gaussian attributes are replaced by
codebook entries $q_{z_j}$,
\begin{equation}
\theta_j \approx q_{z_j},
\qquad
z_j\in\{1,\ldots,K\},
\end{equation}
the storage changes from approximately $\mathcal{O}(Nd)$ parameters to a
codebook plus $N$ indices. In our shared-baseline measurement this reduces
model size from 75.50 to 5.54\,MiB, a 92.66\% reduction, while retaining
25.08\,dB PSNR. Their dominant compression mechanisms act on different terms:
PUP primarily reduces $N$, whereas Compact3D primarily reduces $b_g$ through vector quantization.
}

{\color{red}
The native results in \cref{tab:efficiency_native} show why smaller
representations do not automatically imply faster rendering. The pruned PUP
model contains only about one tenth of the baseline Gaussians, yet its measured
rFPS is lower than the vanilla 3D-GS baseline (155.15 versus 193.46). Hence
$t_{\mathrm{render}}$ is not determined by $N$ alone; visibility, projected
footprint, memory access, and rasterizer implementation remain important.
Octree-GS~\cite{streaminglod} addresses this through view-dependent LoD,
selecting only an appropriate subset of a hierarchical Gaussian representation,
while GoDe~\cite{Sario2025GoDeGOLOD} organizes progressive levels of detail for
scalable compression and transmission,
while gsplat~\cite{ye2025gsplat} primarily improves the rasterization and training
backend rather than changing the scene representation itself. SpeedE3DGS~
\cite{TuYing2025speede3dgs} reduces dynamic-scene cost differently again:
temporal sensitivity pruning removes weakly contributing Gaussians and GroupFlow
shares motion prediction among trajectory clusters, replacing independent
per-Gaussian deformation inference with group-level transformations. Its measured
762.25 rFPS demonstrates the potential of this strategy, but the D-NeRF result is
not directly comparable with the N3DV rows.
}

\subsection{Streaming and Immersive Systems}

\begin{table}[t]
\centering
\caption{\textcolor{red}{Native evaluation of streaming and immersive Gaussian methods on
N3DV \textit{flame\_steak}~\cite{neu3dv}. IGS and 3DGStream measure online
transition updates, whereas GIFStream uses offline two-frame GOP optimization.}}
\label{tab:streaming_native}

\small
\setlength{\tabcolsep}{1.95pt}
\renewcommand{\arraystretch}{1.12}

\begin{tabular}{
>{\raggedright\arraybackslash}p{2cm}
>{\raggedright\arraybackslash}p{1.45cm}
>{\raggedright\arraybackslash}p{1.90cm}
c
>{\raggedright\arraybackslash}p{1.35cm}
c
c
}
\hline
\textbf{Method}
& \textbf{Evaluation unit}
& \textbf{Schedule}
& \textbf{PSNR $\uparrow$}
& \textbf{Operation time $\downarrow$}
& \textbf{Transition/s $\uparrow$}
& \textbf{rFPS $\uparrow$} \\
\hline

IGS~\cite{yan2025instant}
& 1 transition
& 100 refinement iter.
& 33.72
& 7.41 s
& 0.116
& 126.74 \\

3DGStream~\cite{sun20243dgstream}
& 37 transitions
& 15k init.; 150 + 100/update
& 27.34 / 28.18
& 12.63 s/update
& 0.079
& -- \\

GIFStream~\cite{li2025gifstream}
& 2-frame GOP
& 30k
& 28.90
& 2023.62 s/GOP
& --
& 89.75 \\

\hline
\end{tabular}

\vspace{1mm}

\footnotesize
For 3DGStream, PSNR values correspond to stage 1 and stage 2, respectively.
GIFStream is an offline optimization result and is therefore not assigned a transition-update rate.
\end{table}

{\color{red}
Streaming methods replace full per-frame reconstruction with a state transition
\begin{equation}
\mathcal{G}_{t}
=
\mathcal{U}_{\theta}
\!\left(\mathcal{G}_{t-1},I_t\right),
\end{equation}
so practical throughput is governed by the cost of $\mathcal{U}_{\theta}$ rather
than rasterization alone. IGS~\cite{yan2025instant} reduces this cost with an
anchor-driven motion network: multi-view motion evidence is lifted into 3D and
used to predict the motion of the existing Gaussians, while selected key frames
are refined to limit temporal error accumulation. 3DGStream~\cite{sun20243dgstream}
instead represents the transition approximately as
\begin{equation}
\mathcal{G}_{t}
=
\mathcal{T}_{t}(\mathcal{G}_{t-1})
\cup
\mathcal{A}_{t},
\end{equation}
where a compact Neural Transformation Cache predicts Gaussian translation and
rotation and $\mathcal{A}_{t}$ adds Gaussians for newly appearing content.
Thus, IGS emphasizes learned motion transfer with periodic correction, whereas
3DGStream performs explicit per-frame transformation and structural augmentation.
}

{\color{red}
The measurements in \cref{tab:streaming_native} show that this update operator,
not rendering, is the dominant systems bottleneck. IGS reaches 126.74 rFPS after
the representation is updated, but its measured transition rate is only
0.116\,s$^{-1}$ because the native schedule includes 100 refinement iterations
with densification. 3DGStream is slower at 0.079 transitions/s under its
150-stage-1 + 100-stage-2 schedule; the second stage raises mean PSNR from
27.34 to 28.18\,dB, showing a direct quality--update-cost trade-off inside the
streaming loop. Hence
\begin{equation}
t_{\mathrm{stream}}
\approx
t_{\mathrm{motion}}
+
t_{\mathrm{refine}}
+
t_{\mathrm{structure}}
\gg
t_{\mathrm{render}},
\end{equation}
for the measured online systems. Render FPS therefore measures playback capacity,
not the rate at which a new dynamic state can be incorporated.
}

{\color{red}
GIFStream~\cite{li2025gifstream} shifts the objective from rapid online state
transition toward compact transmission of a time-varying Gaussian representation. GIFStream couples reconstruction with compression-aware optimization,
which can be summarized as
\begin{equation}
\mathcal{L}
=
\mathcal{L}_{\mathrm{photo}}
+
\lambda_e\mathcal{L}_{\mathrm{entropy}}
+
\lambda_r\mathcal{L}_{\mathrm{reg}},
\end{equation}
with
\begin{equation}
\mathcal{L}_{\mathrm{entropy}}
\approx
-\sum_t \log
p_{\theta}
\!\left(
\bar V_t \mid \bar V_{<t}
\right).
\end{equation}
It uses a canonical Gaussian space with a deformation field conditioned by
time-dependent feature streams and applies temporal and spatial compression. This explains why its result in
\cref{tab:streaming_native} should not be converted into transition throughput:
the measured 2023.62\,s corresponds to offline optimization of a two-frame GOP,
while the resulting representation still renders at 89.75 rFPS. Streaming
therefore spans two different system problems: minimizing online update latency
and minimizing transmitted temporal representation cost.
Rate-aware and predictive temporal codecs such as 4DGC and P-4DGS further target
this second problem by exploiting inter-frame redundancy in Gaussian streams
\cite{Hu20254DGCR4temporal1,Wang2025P4DGSP4temporal2}; their compression gains
must therefore be assessed together with decoder latency and temporal artifacts.
}

\subsection{Gaussian Splatting SLAM}
\begin{table}[t] \centering \caption{\textcolor{red}{Native evaluation of Gaussian SLAM systems. MonoGS and RTG-SLAM use Replica \textit{office0}~\cite{straub2019replica}, while WildGS-SLAM uses a monocular Crowd sequence. Processed-frame throughput is measured over the complete wrapped run and should not be
interpreted as a controlled ranking across methods.}} \label{tab:slam_native} \small \setlength{\tabcolsep}{2.5pt} \renewcommand{\arraystretch}{1.12} \begin{tabular}{ >{\raggedright\arraybackslash}p{1.85cm} >{\raggedright\arraybackslash}p{2.15cm} >{\centering\arraybackslash}p{0.70cm} >{\centering\arraybackslash}p{1.95cm} >{\centering\arraybackslash}p{1.05cm} >{\centering\arraybackslash}p{1.55cm} >{\centering\arraybackslash}p{1.10cm} } \hline \textbf{Method} & \textbf{Input} & \textbf{Frames} & \textbf{Wrapped time (ms/frame) $\downarrow$} & \textbf{Frames/s $\uparrow$} & \textbf{Trajectory error (m) $\downarrow$} & \textbf{Model size (MiB)} \\ \hline MonoGS~\cite{matsuki2024gaussian} & Replica office0, RGB-D & 12 & 2250.57 & 0.444 & 0.000082 & 2.14 \\ RTG-SLAM~\cite{peng2024rtg} & Replica office0, RGB-D & 20 & 884.60 & 1.130 & 0.00910 & 10.35 \\ WildGS-SLAM~\cite{zheng2025wildgs} & Crowd, monocular & 100 & 1525.64 & 0.655 & 0.00665 & 9.78 \\ \hline \end{tabular} \vspace{1mm} \footnotesize MonoGS and RTG-SLAM report ATE on Replica. WildGS-SLAM reports full-trajectory RMSE; its keyframe RMSE is 0.00371\,m. The wrapped timing includes preprocessing, pose estimation, mapping/update, rendering, and method-specific finalization stages. \end{table}
{\color{red}
Gaussian SLAM couples camera estimation with Gaussian-map optimization.
DyGS-SLAM~\cite{dygs} extends this coupled objective to dynamic-scene localization
and Gaussian reconstruction, illustrating why motion handling must be considered
part of the online estimation loop rather than only a rendering-stage concern.
MonoGS~\cite{matsuki2024gaussian} restricts this joint problem to a selected
set of keyframes,
\begin{equation}
\mathcal{W}=W_k\cup W_r,
\qquad
\min_{\{T_k\},\,\mathcal{G}}
\sum_{k\in\mathcal{W}}
\left(
\lambda_{\mathrm{pho}}E_{\mathrm{pho}}^k
+
\lambda_{\mathrm{geo}}E_{\mathrm{geo}}^k
\right)
+
\lambda_{\mathrm{iso}}E_{\mathrm{iso}},
\end{equation}
where $W_k$ is the covisible local window and $W_r$ contains sampled past
keyframes. Its mapping problem is
therefore locally coupled: increasing the keyframe support improves geometric
constraint but also increases repeated rendering and optimization. This is consistent
with \cref{tab:slam_native}, where the measured short Replica run obtains very low
trajectory error but requires 2250.57\,ms per processed frame. The result demonstrates low trajectory error on this short native execution,
but should not be interpreted as a throughput or accuracy ranking.
}
{\color{red}
RTG-SLAM~\cite{peng2024rtg} reduces this cost by partitioning the Gaussian map as
\begin{equation}
\mathcal{G}
=
\mathcal{G}_{\mathrm{stable}}
\cup
\mathcal{G}_{\mathrm{unstable}},
\qquad
\min_{\mathcal{G}}
\mathcal{L}
\;\longrightarrow\;
\min_{\mathcal{G}_{\mathrm{unstable}}}
\mathcal{L}.
\end{equation}
Stable Gaussians that already explain previous RGB-D observations are largely frozen,
while optimization and rendering concentrate on unstable Gaussians and pixels with
new observations or large color/depth residuals. This converts map maintenance from
a dense update over $N$ primitives toward an active-set problem with
$N_{\mathrm{active}}\ll N$. Its 884.60\,ms wrapped time and 1.130 frames/s in
\cref{tab:slam_native} are consistent with this design goal, although the value is not
directly comparable with MonoGS because the wrapped runs differ in frame count and
stage composition. The important systems result is therefore the reduction of the
optimization domain, not the numerical FPS ordering.
}
{\color{red}
WildGS-SLAM~\cite{zheng2025wildgs} addresses a different failure mode by making
the residual itself uncertainty-aware. Its mapping objective effectively weights
photometric and geometric errors as
\begin{equation}
\mathcal{L}_{\mathrm{map}}
=
\sum_{\mathbf{u}}
\frac{
\lambda_I\mathcal{L}_{I}(\mathbf{u})
+
\lambda_D\mathcal{L}_{D}(\mathbf{u})
}{
\beta(\mathbf{u})^{2}
}
+
\lambda_{\mathrm{iso}}\mathcal{L}_{\mathrm{iso}},
\end{equation}
where larger uncertainty $\beta(\mathbf{u})$ suppresses observations likely to arise
from dynamic distractors. This changes robustness mathematically from treating all
pixels as equally valid map constraints to weighted estimation. In
\cref{tab:slam_native}, WildGS-SLAM processes the longer monocular Crowd sequence
at 1525.64\,ms/frame and obtains a full-trajectory RMSE of 0.00665\,m
(0.00371\,m on keyframes). These values cannot be ranked against the Replica RGB-D
results, but they show that uncertainty-aware monocular estimation remains computationally
substantial even though it operates under a different dynamic monocular input regime. Thus, the three systems reduce different dimensions of the joint SLAM problem:
MonoGS restricts observation support, RTG-SLAM restricts active map variables,
and WildGS-SLAM restricts the influence of unreliable residuals.
}

\subsection{Cross-Family Trade-offs}

{\color{red}
Across the five families, the main distinction is not the Gaussian renderer but
which term dominates the task latency. Using the notation of \cref{sec:realtime},
a cross-family decomposition can be written as
\begin{equation}
t_{\mathcal{T}}
=
t_{\mathrm{geom}}
+
t_{\mathrm{pose}}
+
t_{\mathrm{deform/update}}
+
t_{\mathrm{refine}}
+
t_{\mathrm{render}},
\end{equation}
where only task-relevant terms are active. In single-shot reconstruction,
$t_{\mathrm{geom}}$ dominates: \cref{tab:single_shot_controlled} shows that
decoder-only rFPS is much larger than scene throughput because Gaussian
construction is more expensive than rasterization. Dynamic methods such as
4DGaussians~\cite{4dgs_og} and Deformable-3D-Gaussians~
\cite{Yang2023Deformable3G} shift cost toward temporal deformation and
per-sequence optimization. Streaming systems expose the same separation more
strongly: in \cref{tab:streaming_native}, IGS~\cite{yan2025instant} and
3DGStream~\cite{sun20243dgstream} render an updated representation quickly, while
the transition itself remains orders of magnitude slower. Gaussian SLAM adds
$t_{\mathrm{pose}}$ and map maintenance to this loop, making tracking and mapping,
rather than rasterization, the dominant full-system cost in
\cref{tab:slam_native}.
}

{\color{red}
Efficiency methods operate on a different axis. For
$B(\mathcal{G})\approx Nb_g$, pruning reduces $N$, quantization reduces $b_g$,
and LoD reduces the view-dependent active set $N_{\mathrm{vis}}$, but none of
these implies a proportional decrease in $t_{\mathcal{T}}$. The shared baseline
in \cref{tab:compression_shared} makes this explicit: PUP 3D-GS~
\cite{HansonTuPUP3DGS} and Compact3D~\cite{navaneet2024compact3d} reduce
serialized size by approximately 90\% and 92.66\%, respectively, while the
pruned PUP model does not render faster than the vanilla baseline in
\cref{tab:efficiency_native}. Memory is therefore an independent deployment
constraint rather than a proxy for speed; the failed MotionGS~
\cite{zhu2024motiongs} execution on an 80\,GB A100 provides the complementary
case where optimization becomes infeasible despite the efficiency of Gaussian
rasterization. A useful cross-family operating point is consequently
\begin{equation}
\mathcal{P}
=
\left(
Q,\,
t_{\mathcal{T}},\,
B(\mathcal{G}),\,
t_{\mathrm{render}}
\right),
\end{equation}
and methods should be compared through the resulting trade-off rather than a
single universal FPS or SPS value.
}

{\color{red}
These differences lead to distinct deployment regimes. Feed-forward
single-shot methods are the strongest fit for rapid initialization and
interactive preview because they minimize scene-construction latency, although
the controlled results show a direct quality--latency--memory trade-off.
Per-sequence dynamic methods are better suited when temporal reconstruction
quality is more important than initialization time, since interactive rendering
can be achieved after a substantially heavier optimization stage. Compression
and LoD methods are most useful when storage, bandwidth, or rendering load is
the limiting resource. Streaming methods directly target temporal scene
delivery, but the measured update rates show that interactive playback and
real-time scene refresh remain different system requirements. Finally, Gaussian
SLAM is the natural family for robotics and online mapping because pose
estimation and scene reconstruction are coupled, but the wrapped rates in
\cref{tab:slam_native} show that joint tracking--mapping latency remains the
principal obstacle to video-rate operation. Thus, ``real-time'' Gaussian
Splatting is not a single operating point: the appropriate method depends on
whether the application must optimize reconstruction latency, temporal update
latency, map maintenance, memory/bandwidth, or render-only throughput.
}

\section{Open Problems and Future Directions}
\label{sec:open}
We summarize open problems across the major components of a GS reconstruction pipeline (see \Cref{fig:prelim}), focusing on what blocks unified end-to-end real-time systems.

\subsection{Reporting vs. Reality: Accepted Gaps}
\textbf{Open problems.} ``Real-time'' is often reported at the \emph{stage level} rather than end-to-end, making cross-paper comparisons fragile. Some works report only submodule speed (e.g., VGGT-style lifting \cite{wang2025vggt}), while others report FPS under narrow settings (few views, low resolution) that do not reflect realistic reconstruction workloads \cite{Xu2024DepthSplatCG,Chen2024MVSplatE3}. Depth Anything 3 \cite{lin2025depthanything3recoveringvisual} similarly reports per-image throughput, which does not translate to per-scene throughput for multi-frame/view fusion.

{\color{red}
\textbf{Future directions.} Standardize task-specific speed--quality reporting with explicit input window size, unit of work, timing boundary, hardware, and included pipeline stages.
}

\subsection{Single-Shot Generators: The Deployment Gap}
\textbf{Open problems.} Single-shot GS generators combine heavy backbones with multi-view fusion and sometimes refinement, making ``speed'' hard to interpret. Across depth-conditioned lifting (DepthSplat \cite{Xu2024DepthSplatCG}), sparse-view fusion (MVSplat \cite{Chen2024MVSplatE3}), and learned priors (Splatt3r \cite{smart2024splatt3r}, VGGT-style lifting \cite{wang2025vggt}), runtime often shifts into feature extraction/fusion that is inconsistently included in FPS reports. InstantSplat \cite{fan2025instantsplatsparseviewgaussiansplatting} and pose-free lines \cite{nopo,Hong2024PF3splatPF} further depend on iteration budgets and convergence behavior,
{\color{red}
while custom CUDA or non-standard rasterization/fusion components (e.g., \cite{ye2025gsplat,hanson2025speedysplat}) can complicate deployment and reproducible timing.
}

{\color{red}
\textbf{Future directions.} (i) \emph{Co-design for deployment:} prefer architectures that map cleanly to standard inference graphs with predictable memory/batching. (ii) \emph{Pruning/LoD as first-class:} predict compact Gaussians directly (importance-aware outputs) rather than generating dense sets and pruning afterward.
}

\subsection{Pruning and Efficiency: An Afterthought}
\textbf{Open problems.} Many efficiency levers are applied \emph{after} dense construction, so expensive lifting/fusion/optimization is still paid at full cost. Pruning methods remove large fractions post hoc \cite{HansonTuPUP3DGS,TuYing2025speede3dgs}; quantization/packing reduce bandwidth/storage \cite{niedermayr2024compressedquantizedgs,papantonakis2024reducingpackedsh}; and LoD scales playback \cite{Windisch2025ALODGS,streaminglod}. These benefits often arrive only after dense construction, and view-adaptive pruning can introduce popping/flicker without temporal stabilization \cite{viewadap1,viewadap2}.

{\color{red}
\textbf{Future directions.} (i) \emph{Sparse by construction:} design generators and online updaters that output compact Gaussians natively. (ii) \emph{Temporally stable budgets:} learn selection/budget policies that avoid popping and flicker under fast camera motion.
}

\subsection{Dynamic Streaming: Long-Horizon Stability}
\textbf{Open problems.} Long-horizon streaming remains fragile because dynamics introduce \emph{identity} and \emph{lifecycle} issues: Gaussians must remain consistent under occlusion/disocclusion, topology change, and re-observation after long gaps. Spawnable/time-varying primitives (e.g., FreeTimeGS \cite{wang2025freetimegs}) improve flexibility but exacerbate growth control and bookkeeping. System-wise, streaming pipelines must meet bounded update budgets and avoid latency spikes under cache misses or viewpoint jumps \cite{sun20243dgstream,li2025gifstream,yan2025instant}; many pipelines still resemble offline optimization with real-time playback rather than truly online capture-to-render reconstruction \cite{4dgs_og,Yang2023Deformable3G}.

{\color{red}
\textbf{Future directions.} (i) \emph{Lifecycle + map maintenance:} explicit birth/death, merge/split, and re-anchoring policies to stabilize identity and bound growth \cite{wang2025freetimegs}. (ii) \emph{Canonical + residual streaming:} compact canonical models plus time-dependent deltas \cite{li2025gifstream}. (iii) \emph{Cache-robust updates:} schedules/caches that degrade gracefully under viewpoint jumps \cite{sun20243dgstream,yan2025instant}.
}

\subsection{Rendering: Differentiability}
\textbf{Open problems.} Differentiable rasterization enables end-to-end optimization \cite{3dgs_og}, but full gradients increase memory/overhead and complicate bounded-latency streaming or online updates. Faster rasterizers can prioritize throughput, yet approximations in compositing/sorting/visibility may introduce haloing, incorrect occlusion, and temporal flicker, especially in dense/dynamic regions \cite{ye2025gsplat,hanson2025speedysplat}. Recent work on compositing correctness (e.g., volumetrically consistent formulations) suggests some artifacts are rooted in the rendering model itself \cite{talegaonkar2025volumetrically}.

{\color{red}
\textbf{Future directions.} (i) \emph{Train--deploy separation:} train with differentiable or surrogate rendering, deploy with faster backends, and control mismatch \cite{ye2025gsplat,hanson2025speedysplat}. (ii) \emph{Selective differentiability:} keep full gradients only for keyframes or uncertain regions. (iii) \emph{Correctness complements efficiency:} improved compositing/visibility can reduce artifacts without inflating $N$ \cite{talegaonkar2025volumetrically}.
}

\subsection{SLAM with Gaussians: The Missing Pieces}
\textbf{Open problems.} GS-based SLAM must satisfy tracking deadlines and maintain a map that is both renderable and usable for localization. While recent systems show real-time tracking/mapping \cite{yan2024gs,matsuki2024gaussian,peng2024rtg}, long-term consistency (loop closure/relocalization), bounded growth via maintenance, and robustness to dynamic-object corruption remain under-emphasized in in-the-wild settings \cite{zheng2025wildgs}.

{\color{red}
\textbf{Future directions.} (i) \emph{Loop-consistent Gaussian maps:} loop-aware merge/re-anchor/de-duplicate updates to preserve render quality and pose consistency. (ii) \emph{Maintenance-first SLAM:} pruning, keyframe selection, and local submap management as core components to keep latency stable \cite{peng2024rtg}. (iii) \emph{Dynamic-robust mapping:} lightweight masking/segmentation and static/dynamic separation to prevent permanent map pollution \cite{zheng2025wildgs}.
}
\section{Conclusion}
This survey reviewed Gaussian Splatting (GS) for real-time scene reconstruction. Rather than only listing methods, we organized prior work into a pipeline-centric taxonomy and highlighted the trade-offs that determine whether a method is merely \emph{real-time in a submodule} or deployable end-to-end.

\textbf{Strengths and value to the field.}
The main strength of this work is its \emph{synthesis across components}. We connect design choices (e.g., temporal parameterization vs.\ stability, pruning/quantization/LoD vs.\ fidelity, and rasterization choices vs.\ artifacts) to system-level outcomes such as latency, memory growth, robustness to pose/depth noise, and long-horizon stability. {\color{red}
The taxonomy, comparison framework, and failure-mode analysis help researchers position new contributions and practitioners reason about capture-to-render pipelines.
}

\textbf{Weaknesses and limitations.}
Our taxonomy necessarily simplifies a fast-moving hybrid space, where many systems combine amortized initialization with refinement or mix static and dynamic representations. Comparability is also limited: papers report speed and memory at different stages (e.g., per-image FPS vs.\ per-scene throughput), under different input-window sizes and hardware assumptions. Thus, end-to-end conclusions about ``real-time'' readiness should be read as pipeline-aware guidance rather than a standardized benchmark ranking.

\textbf{Future work.}
{\color{red}
The main remaining directions are task-specific standardized evaluation, deployment-aware scene generation, sparse-by-construction representations, long-horizon dynamic and streaming maintenance, and robust Gaussian SLAM. Together, these problems shift the focus from fast rasterization toward maintaining quality, bounded latency, and manageable state throughout the reconstruction pipeline.
}

\section*{CRediT Authorship Contribution Statement}
Ashim Dahal: Writing – original draft,  Methodology, Formal analysis, Conceptualization; Rabab Abdelfattah: Writing – review \& editing, Validation, Supervision; Nick Rahimi: Writing – review \& editing, Validation, Supervision.

\noindent\textbf{Acknowledgments.} The authors used a large language model (LLM) solely for English grammar, clarity, and \LaTeX{} formatting assistance; all technical content, interpretations, opinions, and conclusions are the authors' responsibility.
\section*{Code and Data Availability}
{\color{red}
Benchmark and evaluation code will be made publicly available upon acceptance.
The datasets used in the evaluation are publicly available from their respective
sources.
}
\bibliography{main}
\bibliographystyle{elsarticle-num}

\end{document}